\documentclass[letterpaper,11pt]{article}
\usepackage[utf8]{inputenc}
\usepackage[T1]{fontenc}

\usepackage[english]{babel}
\usepackage{authblk}
\usepackage[margin=1in]{geometry}

\usepackage{amsfonts}       
\usepackage{nicefrac}       
\usepackage{microtype}      
\usepackage{amsmath,bbm,bm,mathtools} 
\usepackage{float}
\usepackage{amssymb}
\usepackage{amsthm}
\usepackage{algorithm}
\usepackage{algorithmic}
\usepackage{booktabs}       
\usepackage{graphicx}
\usepackage{multirow}
\usepackage{subcaption}
\usepackage[colorlinks=true, allcolors=black]{hyperref}
\usepackage{xcolor}
\usepackage{comment}
\usepackage[shortlabels]{enumitem}
\usepackage[normalem]{ulem}

\usepackage{natbib}
\setcitestyle{authoryear,open={(},close={)}}

\newtheorem{theorem}{Theorem}
\newtheorem{corollary}{Corollary}

\theoremstyle{plain}

\def\eqref#1{equation~\ref{#1}}

\def\1{\bm{1}}

\def\vone{{\mathbbm{1}}}

\DeclareMathAlphabet{\mathsfit}{\encodingdefault}{\sfdefault}{m}{sl}
\SetMathAlphabet{\mathsfit}{bold}{\encodingdefault}{\sfdefault}{bx}{n}

\def\gC{{\mathcal{C}}}

\def\gF{{\mathcal{F}}}
\def\gG{{\mathcal{G}}}

\def\gX{{\mathcal{X}}}
\def\gY{{\mathcal{Y}}}

\def\sR{{\mathbb{R}}}

\newcommand{\E}{\mathbb{E}}

\newcommand{\R}{\mathbb{R}}

\DeclareMathOperator*{\argmax}{arg\,max}
\DeclareMathOperator*{\argmin}{arg\,min}

\definecolor{DarkGreen}{rgb}{0.1,0.5,0.1}
\definecolor{DarkRed}{rgb}{0.5,0.1,0.1}
\definecolor{DarkBlue}{rgb}{0.1,0.1,0.5}
\definecolor{Gray}{rgb}{0.2,0.2,0.2}

\def\standard{\textsc{standard}}

\def\standard{\textsc{standard}}
\def\naive{\textsc{naive}}
\def\conditional{\textsc{conditional}}
\def\condnaive{\textsc{conditional \scriptsize{(NAIVE)}}}
\def\groupwise{\textsc{groupwise}}

\renewcommand{\P}{\mathbb{P}} 

\makeatletter
\newcommand\footnoteref[1]{\protected@xdef\@thefnmark{\ref{#1}}\@footnotemark}
\makeatother

\title{\Large{Conformal Prediction Sets with Improved Conditional Coverage \\using Trust Scores}}

\author{\large{Jivat Neet Kaur$^*$ ~~~~ Michael I. Jordan$^{*\dagger}$ ~~~~ Ahmed Alaa$^*$}}
\affil{$^*$University of California, Berkeley \\
$^\dagger$Inria, Paris}

\date{}

\begin{document}

\maketitle



  
  

 

  

\begin{abstract}
Standard conformal prediction offers a marginal guarantee on~coverage,~but~for prediction sets to be truly useful, they should ideally ensure coverage {\it conditional}~on each test point. Unfortunately, it is impossible to achieve exact, distribution-free conditional coverage in finite samples. In this work, we propose an alternative~conformal prediction~algorithm that targets coverage where it matters most---in instances where a~classifier~is {\it overconfident in its incorrect predictions}. We start by dissecting miscoverage events in marginally-valid conformal prediction, and show that miscoverage rates vary based on~the~classifier's confidence and its deviation from the Bayes optimal classifier. Motivated by this insight, we develop a variant of conformal prediction that targets coverage conditional on a reduced set of two variables: the classifier's confidence in a prediction~and~a~nonparametric {\it trust score} that measures its deviation from the Bayes classifier. Empirical evaluation on multiple image datasets shows that our method generally improves conditional coverage properties compared to standard conformal prediction, including class-conditional coverage, coverage over arbitrary subgroups, and coverage~over~demographic~groups.
\end{abstract}

\section{Introduction}

Machine learning models are envisioned to inform decision-making in high-stakes applications such as medical diagnosis \citep{a35d222dd67c493cb31641fc3d14d698, ELFANAGELY2021346, Caruana2015IntelligibleMF}. Consequently, there is a critical need for actionable and useful uncertainty estimates to mitigate the risks associated with incorrect decisions influenced by overconfident models. Conformal prediction~\citep{vovk2005algorithmic} is a framework for constructing prediction sets that provide finite-sample \textit{marginal coverage} guarantees without making any modeling or distributional assumptions beyond exchangeability; i.e., the sets contain the correct output with a specified probability. Given a calibration dataset $\{(X_i,Y_i)\}_{i=1}^{n}$ and a new test point $(X_{n+1}, Y_{n+1})$, split conformal prediction (referred to as simply ``conformal prediction'' from here on) constructs a prediction set $\gC(X_{n+1}) \subseteq \gY$ that satisfies,
\begin{align}
\label{eq:split_marginal_cov}
\P(Y_{n + 1} \in \gC(X_{n + 1})) \geq 1 - \alpha,
\end{align}
for $\alpha \in (0, 1)$. Such marginal validity does not, however, ensure~that~the~prediction~sets~are~actionable in arbitrary contexts, as coverage can be unacceptably~poor~for~individual~predictions.

To highlight the limitations of marginal validity, consider a scenario where a model~is~used~to~diagnose a disease in a population where 90\% of the cases are straightforward to diagnose,~while~10\%~are challenging. Prediction sets that only cover in the straightforward cases may achieve 90\% marginal coverage, but may be useless for the cases where uncertainty estimates are most critical for clinicians. Ideally, one would like to construct prediction sets that satisfy \textit{conditional coverage}; i.e., $\P(Y_{n + 1} \in \gC(X_{n + 1})|X_{n+1}=x) \geq 1 - \alpha, \forall x$. Unfortunately, however, it is well known that it is impossible to attain distribution-free exact conditional coverage in a meaningful sense \citep{lei2014distribution, vovk2012conditional, barber2021limits}. The question becomes one of approximate conditional coverage.

\begin{figure}[t] 
    \centering
    \includegraphics[width=1\linewidth]{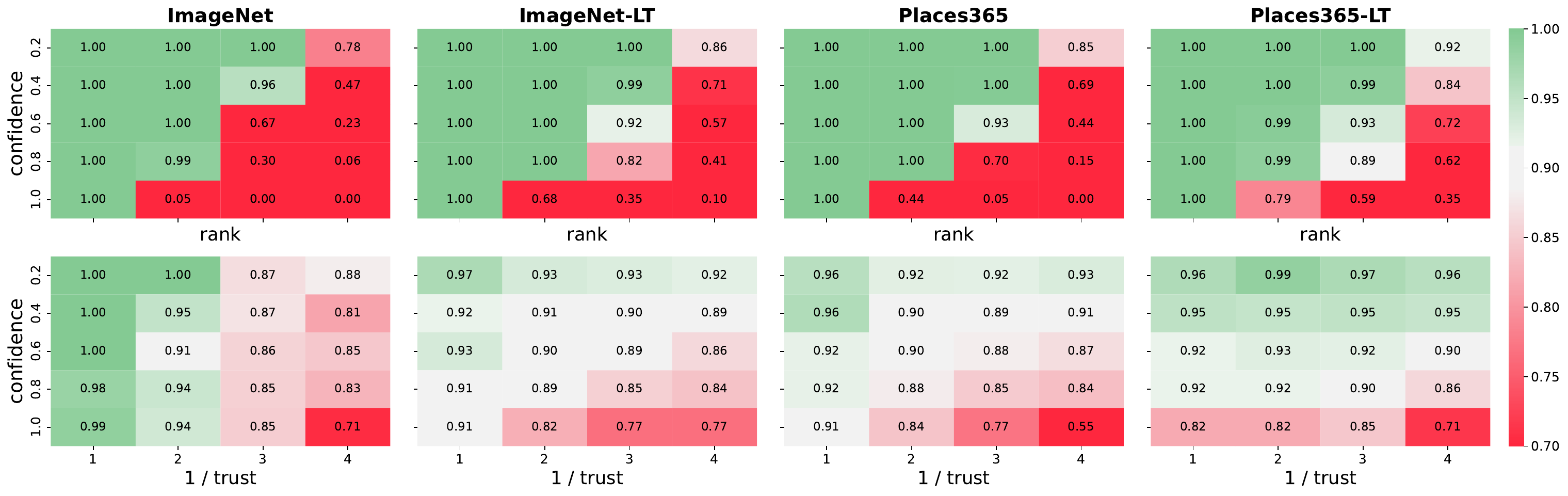} 
    \caption{\textbf{Miscoverage patterns in standard conformal prediction.} Illustration of conditional coverage of standard conformal prediction ($\standard$) over regions of the feature space binned by model confidence and rank of the true class (top) ($\mbox{\textup{\fontsize{10.5pt}{12.2pt}\textsf{Conf}}} \times \mbox{\textup{\fontsize{10.5pt}{12.2pt}\textsf{Rank}}}$), and confidence and \textit{trust score} (bottom) ($\mbox{\textup{\fontsize{10.5pt}{12.2pt}\textsf{Conf}}} \times \mbox{\textup{\fontsize{10.5pt}{12.2pt}\textsf{Trust}}}$). We set $\alpha = 0.1$; hence, red bins indicate undercoverage (coverage $<$ 0.9) and green bins indicate overcoverage (coverage $>$ 0.9). We split samples into equal-size bins based on rank and trust score. As a special case for ImageNet, we manually define the rank bins, as $\sim$75\% test samples have accurate predictions. Models are calibrated using temperature scaling.}
    \label{fig:mainfig}
\end{figure}

In this work, we introduce a relaxed objective for conditional coverage and a variant of conformal prediction that ensures that prediction sets adapt to the true uncertainty of test instances.~Since~the~feature space $\mathcal{X} \subseteq \mathbb{R}^d$ can be high-dimensional, achieving approximate $X$-conditional coverage is challenging. \cite{gibbs2023conformal} proposed a method that can theoretically achieve conditional coverage with respect to any function class; however, implementing this with a function class that~corresponds~to $X$-conditional coverage is infeasible. Our key idea is to instead identify a lower-dimensional variable $V$, where the value of $V$ is indicative of whether standard conformal prediction will over- or under-cover the corresponding test instance. Based on $V$, we propose a function class that is both practical to implement and yields conditional guarantees important for decision-making. We then construct prediction sets that achieve (approximate) conditional coverage with respect to $V$, i.e., 
\begin{align}
\label{eq:v_cond_cov}
    \P(Y_{n + 1} \in \gC(X_{n + 1})|V_{n+1}=v) \geq 1 - \alpha, \, \forall v.
\end{align}
Our choice of the variable $V$ is driven by an analysis of how miscoverage events are distributed across test instances in standard conformal prediction. In the classification setting, we select $V$~as~a~statistic that identifies instances where conformal prediction is most likely to fail---specifically when~the~classifier is {\it overconfident in its incorrect predictions}. Figure \ref{fig:mainfig} (top) illustrates the relationship~between~miscoverage rates in standard conformal prediction and the classifier's overconfidence as measured by its reported confidence (top softmax output) and the rank of $Y_{n+1}$ in its sorted softmax~probabilities. The miscoverage patterns in Figure \ref{fig:mainfig} suggest that these two factors are predictive of whether conformal prediction over- or under-covers a test instance. 

A variant of conformal prediction that achieves the relaxed conditional guarantee in (\ref{eq:v_cond_cov}) with respect to the statistic $V = \{\mbox{\textup{\textsf{Conf}}}, \mbox{\textup{\textsf{Rank}}}\}$---where \mbox{\textup{\textsf{Conf}}} is the classifier's confidence and \mbox{\textup{\textsf{Rank}}} is the rank of $Y_{n+1}$---ensures that the resulting prediction sets cover both low- and high-uncertainty test instances. Such a procedure would produce more balanced coverage patterns compared to standard conformal prediction. Since the \mbox{\textup{\textsf{Rank}}} variable depends on $Y_{n+1}$, which is not available to us during test time, we propose a practical choice for $V$ that assesses the classifier's overconfidence through the model confidence (\mbox{\textup{\textsf{Conf}}}) and a nonparametric \textit{trust score} (\mbox{\textup{\textsf{Trust}}}) that measures the disagreement of the model predictions with the Bayes-optimal classifier \citep{NEURIPS2018_7180cffd} (Section~\ref{sec:trust}).

We perform an extensive evaluation on four large-scale image classification datasets: ImageNet~\citep{russakovsky2015imagenet}, Places365~\citep{zhou2018places}, and their corresponding long-tail versions ImageNet-LT and Places365-LT~\citep{openlongtailrecognition}. Since conditional coverage has not been studied in this setting previously, we propose a suite of evaluation metrics to measure approximate conditional coverage. We find that our proposed method reduces the coverage gap across test instances as evaluated by these metrics in all datasets. We also perform evaluation on the Fitzpatrick 17k dataset~\citep{groh2021evaluating} for skin condition classification in clinical images, where we are able to improve coverage across different skin types without access to type annotations.

\subsection{Related work}

\paragraph{Group-conditional conformal prediction.} A widely adopted relaxation of conditional coverage in prior work is based on group-conditional guarantees of the form $\P(Y_{n + 1} \in \gC(X_{n + 1}) \mid X_{n+1} \in G) \geq 1 - \alpha$ for all groups $G \in \gG$~\citep{fe5374fa2e6b4c269718c6de868bab26,barber2021limits,jung2023batch}. This concept is often motivated by the idea that instead of $X$-conditional coverage, one can ensure the coverage guarantee holds over predetermined subgroups that might otherwise be underserved by marginal coverage~\citep{Romano2020With}. Mondrian conformal prediction~\citep{fe5374fa2e6b4c269718c6de868bab26} achieves exact group-conditional coverage in finite samples when the groups in $\gG$ are disjoint.~The~procedure~involves splitting the calibration data into subgroups and then separately calibrating on each group.~Within~this~framework,~\citet{Romano2020With} focus on achieving equal coverage over disjoint protected groups of interest.~\citet{barber2021limits} propose an approach that allows groups in $\gG$ to overlap; however, this method can be highly conservative and result in wide prediction intervals that over-cover. To provide practical, ``multivalid" coverage guarantees over arbitrary subgroups,~\citet{jung2023batch} propose learning quantile multiaccurate predictors by minimizing the pinball loss over the class of functions $\mathcal{F} = \{\sum_{G \in \mathcal{G}} \beta_G \vone\{x \in G\}: \beta \in \R^{|\mathcal{G}|}\}$. While~\citet{jung2023batch} provide PAC-style guarantees,~\citet{gibbs2023conformal} propose a conditional conformal procedure that yields exact coverage guarantees over  collections of groups in finite samples, and also extends beyond the group setting to finite-dimensional classes. That said, as noted earlier, it is not feasible to run this method with a function class that can guarantee approximate $X$-conditional coverage. Our work is complementary to this line of work: we propose a practical way to construct a function class that guarantees an interpretable notion of conditional coverage broader than group-conditional coverage over pre-specified groups.

We note there are some recent works that share similar motivation of learning features from data to improve conditional coverage. \citet{yuksekgonul2023beyond} propose a density-based \textit{atypicality} notion to improve calibration and conditional coverage with respect to input atypicality. They implement a special case of Mondrian conformal prediction to improve coverage in high atypicality or low confidence groups. In our work, the~goal~is~to improve conditional coverage more generally, and our method differs from variants of Mondrian conformal prediction that cluster the covariate space.~\citet{KiyaniPH24} propose to learn partitioning of the covariate space such that points in the same partition are similar
in terms of their prediction sets in order to improve conditional validity. They present an algorithm that iteratively updates the partitioning and prediction sets over a given calibration data set. While they learn low-dimensional features from the calibration data, we study general patterns of miscoverage in standard conformal prediction and propose a two-dimensional statistic that consistently demonstrates its effectiveness in settings that extend beyond well-defined groups. In the regression setting,~\citet{sesia2021conformal} learn conditional histograms from the data to detect the skewness of $ Y \mid X$ and estimate the quantiles of the conditional distribution. 

\paragraph{Other approaches for improving conditional coverage.} With the goal of achieving approximate $X$-conditional coverage, previous works have proposed new conformity score functions~\citep{romano2019conformalized, romano2020classification,angelopoulos2021uncertainty} that yield significant practical improvements.~\citet{ding2023classconditional} focus on achieving class-conditional coverage instead, and introduce a clustered conformal prediction method for $Y$-conditional coverage. As argued by~\citet{angelopoulos2021uncertainty} and~\citet{ding2023classconditional}, we note that in high-signal problems like image classification where $Y$ is perfectly determined by $X$, $X$-conditional coverage is less interpretable as an objective.

\section{Preliminaries}
\subsection{Standard conformal prediction} 
\label{sec:2.1}
In this paper, we consider a classification setting in which each input feature $X_i \in \gX$ is associated with a class label $Y_i$ drawn from a discrete set of possible classes $\gY$. Let $s : \mathcal{X} \times \mathcal{Y} \to \sR$~be~a~\emph{conformity score} function that measures how well the label $y$ ``conforms" to a model prediction at $x$, where lower scores indicate better agreement. (A simple choice for the score is $s(x, y) = 1-\hat{f}_y(x)$,~where $\hat{f}: \mathcal{X} \to \mathcal{Y}$ is a pretrained classifier and $\hat{f}_y(x)$ is its softmax output for class $y$). Given the calibration data set, $\{(X_i, Y_i)\}_{i=1}^n$, and a model $\hat{f}$, a conformal prediction set $\gC(X_{n +1})$ for a test point $X_{n +1}$ is constructed by evaluating the conformity scores $s_i = s(X_i, Y_i), 1 \leq 1 \leq n$. Then, we compute $\hat{q}$ as the $\lceil (n + 1) (1 - \alpha) \rceil/n$ empirical quantile of $\{s_i\}_{i=1}^n$, and use $\hat{q}$ to construct the prediction sets
\begin{equation}
\label{eq:split_conformal_set}
\gC(X_{n + 1}) = \{y : s(X_{n + 1}, y) \leq \hat{q}\}.
\end{equation}
We refer to this as $\standard$ split conformal prediction following \citet{ding2023classconditional}. $\standard$ conformal prediction guarantees marginal validity as described in (\ref{eq:split_marginal_cov}) as long as the calibration~and~test scores $s_1,\dots,s_n, s_{n+1}$ are exchangeable. However, as discussed earlier, marginal coverage~may~be insufficient for $\gC$ to be practically useful and we aim for a stronger notion of conditional coverage.

\subsection{Dissecting miscoverage patterns in standard conformal prediction}
\label{sec:2.2}

{\bf Exact conditional coverage with an oracle classifier.} Imagine an ``oracle'' classifier $f^*$~which~perfectly~captures the distribution $f^*_y(x) = P(Y = y | X = x), \forall y \in \gY, x \in \gX$. \citet{romano2020classification} showed that one could construct optimal prediction sets $\gC^\text{oracle}(x)$ with exact conditional coverage by leveraging the {\it order statistics} $f^*_{(1)}(x) \geq f^*_{(2)}(x) \geq \dots \geq f^*_{(|\gY|)}(x)$, for $\{f^*_y(x): y \in\gY\}$, as follows:
\begin{equation*} 
\gC^\text{oracle}(x) = \{\text{`$y$' indices of the} \,\, k \,\,\text{largest} \,\, f^*_y(x)\}, \text{where} \,\, k = \inf \left\{ k' : \mbox{$\sum_{j=1}^{k'}$} f^*_{(j)}(x) \geq 1 - \alpha\right\}.
\end{equation*}
With knowledge of the true probabilities $\{f^*_y(x)\}_y$, $k = \inf \{ k' : \mbox{\small $\sum$}_{j=1}^{k'} f^*_{(j)}(x) \geq 1 - \alpha\}$ can be thought of as a generalization of the conditional quantile function for continuous outcomes. In this sense, the oracle sets are conditionally valid because they correspond to the $(1-\alpha)$ quantile of $Y|X$. In practice, however, constructing such sets is impossible because we do not have access to $f^*$. 

\paragraph{Overconfidence and miscoverage patterns for $\standard$.} A common implementation of conformal prediction approximates the oracle algorithm above using $\hat{f}$ as a ``plug-in'' estimate of $f^*$ to~compute~the~score 
\begin{equation} \label{oracle:score}
s(x, y) = \mbox{$\sum_{j=1}^{k}$} \hat{f}_{(j)}(x), \text{where} \,\, \hat{f}_{(k)}(x) = \hat{f}_y(x),
\end{equation}
where $\hat{f}_{(j)}(x)$ denotes the $j^\textup{th}$ sorted value of the softmax outputs of $\hat{f}$, and $k$ is the index in the~sorted~order that corresponds to true class $y$ \citep{romano2020classification, angelopoulos2021uncertainty}. This variant of conformal prediction may empirically improve adaptivity of the resulting sets, but still guarantees only marginal validity in finite samples. This means some regions of the feature space will be over-covered, while others will be under-covered. Intuitively, this variant of $\standard$ is more likely to miscover instances~where~$\hat{f}$ is a poor approximation of the oracle $f^*$, which are also the cases where uncertainty quantification is most critical. 

Let \mbox{\textup{\textsf{Rank}}}$(X_{n+1}, Y_{n+1})$ be the position of softmax output $\hat{f}_{Y_{n+1}}(X_{n+1})$ in the decreasing sorted order of class probabilities $\hat{f}_{(1)}(X_{n + 1}), \dots, \hat{f}_{(|\gY|)}(X_{n + 1})$, and \mbox{\textup{\textsf{Conf}}}$(X_{n+1}) = \max_y \hat{f}_{y}(X_{n+1})$. The variables \mbox{\textup{\textsf{Rank}}}$(x, y)$ and \mbox{\textup{\textsf{Conf}}}$(x)$ measure a model's overconfidence---a model is considered overconfident when it assigns a high value on its largest softmax output while the true class ranks poorly in the descending order of predicted class probabilities. In Figure \ref{fig:mainfig} (top), we visualize empirically the coverage rates of $\standard$ in different strata of \mbox{\textup{\textsf{Conf}}} and \mbox{\textup{\textsf{Rank}}} across various image classification datasets.~As~we~can see, if the model's prediction is incorrect (i.e., \mbox{\textup{\textsf{Rank}}}$(X_{n+1}, Y_{n+1}) > 1$), an increase in overconfidence leads to a reduction in coverage probability.
This monotonic relationship holds consistently across all predictions that share a similar value for \mbox{\textup{\textsf{Rank}}}. Additionally, we observe that for a fixed \mbox{\textup{\textsf{Conf}}}, coverage probability decreases as \mbox{\textup{\textsf{Rank}}} increases. 

This finding suggests that, given the values~of~\mbox{\textup{\textsf{Conf}}} and \mbox{\textup{\textsf{Rank}}} for a test instance, we can determine its position within the strata in Figure \ref{fig:mainfig}, predict whether $\standard$ will likely over- or under-cover the true label, and adjust the conformal prediction set accordingly to achieve more balanced coverage across test instances. These miscoverage patterns are a result of the conformity score intended to approximate the oracle algorithm, and hold across different datasets irrespective of the underlying data distribution. Hence, this finding is expected to generalize across datasets and likely not be alleviated by popular calibration methods such as temperature scaling as seen in Figure \ref{fig:mainfig}.

\section{Conditional conformal prediction with trust scores} 
Consider a scenario where the feature space $\mathcal{X}$ is discrete, and we have access to a large calibration set that includes all possible values in $\mathcal{X}$. In this case, one approach to constructing~conditionally~valid sets is to select a distinct threshold $\hat{q}$ in (\ref{eq:split_conformal_set}) for each $x \in \mathcal{X}$. This procedure would assign a~larger~$\hat{q}$~for instances where the model $\hat{f}$ is more prone to errors, and a smaller $\hat{q}$ where errors are less likely. However, in practice, this procedure is infeasible because $\mathcal{X}$ is typically high-dimensional or continuous. More generally, \cite{lei2014distribution} and \cite{barber2021limits} have shown that distribution-free conditionally valid predictive inference is impossible to attain meaningfully.

The key idea behind our proposed method is to condition on a lower-dimensional statistic $V \in \mathcal{V}$, rather than the full feature space $\mathcal{X}$. We select this statistic as a proxy to identify test instances prone to error, enabling the selection of a distinct threshold $\hat{q}$ for each $v \in \mathcal{V}$, rather than conditioning on each individual point in $\mathcal{X}$. While this procedure does not achieve $X$-conditional coverage, it provides coverage conditioned on the likelihood of under-coverage by $\standard$. The miscoverage patterns discussed in Section \ref{sec:2.2} motivate the selection of~$V$~as~$V=\{\mbox{\textup{\textsf{Conf}}}, \mbox{\textup{\textsf{Rank}}}\}$. Consequently, instead of strict conditional coverage, $\P(Y_{n + 1} \in \gC(X_{n + 1}) \mid X_{n + 1}=x) \geq 1-\alpha,$ for all $x \in \mathcal{X}$, we adopt a more relaxed notion of conditional coverage as follows:
\begin{equation}
    \P(Y_{n + 1} \in \gC(X_{n + 1}) \mid \mbox{\textup{\textsf{Conf}}}(X_{n+1}) = c, \,\mbox{\textup{\textsf{Rank}}}(X_{n+1}, Y_{n+1})=r) \geq 1-\alpha, 
\end{equation}
for all $c \in [0, 1]$ and $r \in \{1,\ldots,|\mathcal{Y}|\}$. However, the label $Y_{n+1}$ is not available at test time, and thus $\mbox{\textup{\textsf{Rank}}}(X_{n+1}, Y_{n+1})$ cannot be used to construct the prediction sets. In the next section, we~propose~an alternative to the \mbox{\textup{\textsf{Rank}}}$(X, Y)$ variable that can be computed using calibration data.

\subsection{Implementation using trust scores}
\label{sec:trust}
The \mbox{\textup{\textsf{Rank}}}$(X, Y)$ variable measures how far from the top softmax score a classifier ranks the~true~class $Y$ for a given input $X$. Since we do not have access to the label $Y$, we use the \textit{trust score} proposed in \citet{NEURIPS2018_7180cffd} as a proxy to approximate the miscoverage patterns characterized by \mbox{\textup{\textsf{Rank}}} (Figure \ref{fig:mainfig}). The trust score is a nonparametric statistic that measures~the agreement between the classifier $\hat{f}$ and the Bayes-optimal classifier on a given testing point $X$. Formally, the trust score \mbox{\textup{\textsf{Trust}}}$(X; \hat{f})$ for a classifier $\hat{f}$ on test point $X$ is defined as 
\begin{align} \label{trust:score}
\mbox{\textup{\textsf{Trust}}}(x; \hat{f}) := d\left(x, \widehat{H_\delta}(P_{\tilde{y}})\right)/d\left(x, \widehat{H_\delta}(P_{\hat{y}})\right),
\end{align}
where $\widehat{H_\delta}(P) := \{ x \in X : r_k(x) \le \varepsilon \}$; $k$-NN radius $r_k(x) := \inf\{ r > 0 : |B(x, r) \cap X| \ge k\}$, $\varepsilon := \inf\{ r > 0 : |\{x \in X : r_k(x) > r\}| \le \delta\cdot n\}$.

The evaluation of \mbox{\textup{\textsf{Trust}}} proceeds in two stages: first,~a~$\delta$\emph{-high-density-set} $\widehat{H_\delta}(P_\ell)$ (for continuous density function $P$ with compact support $\gX$) is estimated for each class $\ell$ from the training data by filtering out a $\delta$-fraction of samples with lowest density. Then, for a given test sample, the trust score ($\mbox{\textup{\textsf{Trust}}}(X$)) (\ref{trust:score}) is computed as the ratio of the Euclidean distance between $X$ and the nearest point in the training set with class label \textit{different} from the top-1 predicted label (say, $\tilde{y}$), and the distance between $X$ and the nearest point with class label \textit{same} as the top-1 predicted label by the classifier (say, 
$\hat{y}$). We provide the implementation details in Appendix~\ref{sec:app_function_class}.

Theorem 4 in \citet{NEURIPS2018_7180cffd} provides the following guarantee for the trust score: for~labeled~data distributions with well-behaved class margins, when the trust score is large, the classifier~likely~agrees with the Bayes optimal classifier $h^*(x) := \argmax_{\ell \in \gY} \P(y = \ell | x)$, and when the trust score is small, the classifier likely disagrees with it. More formally, under certain regularity assumptions, $\mbox{\textup{\textsf{Trust}}}(x; \hat{f})$ satisfies the following with high
probability uniformly over all $x \in \gX$ and all classifiers $\hat{f}: \mathcal{X} \to \mathcal{Y}$ simultaneously for sufficiently large training set:
\begin{align*}
\mbox{\textup{\textsf{Trust}}}(x; \hat{f}) < \gamma &\Rightarrow \hat{f}(x) \neq h^*(x),\\
1/\mbox{\textup{\textsf{Trust}}}(x; \hat{f}) < \gamma &\Rightarrow \hat{f}(x) = h^*(x),
\end{align*}

\noindent for some data-dependent threshold $\gamma$. The paper can be consulted for the full theorem and proof. This result suggests that the $\textsf{Trust}$ score stratifies prediction instances based on their agreement with the Bayes-optimal classifier. Given the Bayes-optimal classifier has low error, high probability of agreement with the Bayes-optimal classifier can help identify correct predictions (``trustworthy" examples), whereas high probability of disagreement can indicate the classifier is making an unreasonable decision. In essence, it provides a coarser analog to the binning in Figure 1 into $\textsf{Rank}=1$ and $\textsf{Rank}>1$ strata. Since the $\textsf{Trust}$ score evaluates this agreement through thresholding, higher continuous values of trust score will likely correspond to improved $\textsf{Rank}$. We present an empirical analysis of the correlation between $\mbox{\textup{\textsf{Trust}}}$ and  $\mbox{\textup{\textsf{Rank}}}$ in Appendix~\ref{sec:app_trust_rank}.

\subsection{Algorithm}
We introduce our $\conditional$ method for conformal prediction with the~goal~of~achieving~approximate conditional coverage over the reduced variable set $V = \{ \mbox{\textup{\textsf{Conf}}}(X_{n+1}), \mbox{\textup{\textsf{Trust}}}(X_{n+1})\}$: 
\begin{align}
\label{eq:V_cond_cov}
\P(Y_{n + 1} \in \gC(X_{n + 1}) \mid \mbox{\textup{\textsf{Conf}}}(X_{n+1})\in I_1, \mbox{\textup{\textsf{Trust}}}(X_{n+1}) \in I_2) \geq 1 - \alpha,
\end{align} where sub-intervals $I_1$ and $I_2$ are some discretization of $[0, 1]$ and $(0, \infty)$ respectively. The exact conditional coverage guarantee $\P(Y_{n + 1} \in \gC(X_{n + 1}) |X_{n+1}=x) = 1 - \alpha, \,\,\forall x \in \mathcal{X}$, is equivalent to a marginal guarantee over all measurable functions $f$:
\begin{align}
\E[f(X_{n + 1}) \cdot (\vone \{Y_{n + 1} \in \gC(X_{n + 1})\} &- (1 - \alpha))] = 0, \quad \text{for all measurable $f$}.
\end{align}
If $f(x) = x \mapsto 1$, we recover marginal coverage. \citet{gibbs2023conformal} propose a relaxation~of~the~exact conditional coverage guarantee over all measurable $f$ to all $f$ belonging to some function class $\gF$,
\begin{align}
\label{eq:cond_cov_objective}
\E[f(X_{n + 1}) \cdot (\vone \{Y_{n + 1} \in \gC(X_{n + 1})\} &- (1 - \alpha))] = 0, \quad \text{for all $f \in \gF$}.
\end{align}
A special case of this guarantee is group-conditional coverage; i.e., $\P (Y_{n + 1} \in \gC(X_{n + 1}) \mid X_{n + 1} \in G) = 1 - \alpha$ for all $G$ belonging to some collection of groups $\gG$, where $\mathcal{F} = \{\sum_{G \in \mathcal{G}} \beta_G \vone\{x \in G\}: \beta \in \R^{|\mathcal{G}|}\}$. The above conditional coverage guarantee can be achieved by fitting an augmented quantile regression problem over $\gF$ where the unknown conformity score $s_{n+1}$ is imputed as $s$. The quantile estimate $\hat{g}_s$ is fit using the \textit{pinball loss} $\ell_{\alpha}(g(X_i),s_i)$ as follows

\vspace{-.1in}
\begin{equation}
\label{eq:finite_dim_reg}
\hat{g}_s = \argmin_{g \in \mathcal{F}} \frac{1}{n+1}\sum_{i=1}^n \ell_{\alpha}(g(X_i),s_i) + \frac{1}{n+1} \ell_{\alpha}(g(X_{n+1}),s),
\end{equation}
where $\ell_{\alpha}(g(X_i),s_i) = (1 - \alpha)(s_i - g(X_i))_{+} + \alpha(g(X_i) - s_i)_{+} $. We compute the prediction set by 
\begin{equation}
\label{eq:finite_dim_pred_set}
\gC(X_{n + 1}) = \{y : s(X_{n+1},y) \leq \hat{g}_{s(X_{n+1},y)}(X_{n+1})\}.
\end{equation}
For a finite-dimensional linear class $\mathcal{F} = \{\Phi(\cdot)^\top\beta : \beta \in \R^d\}$ over the basis $\Phi: \mathcal{X} \to \mathbb{R}^d$,~\citet{gibbs2023conformal} show that we can achieve an upper bound on coverage in (\ref{eq:cond_cov_objective}).

\begin{theorem}[Theorem 2~\citet{gibbs2023conformal}]
\label{thm:finite_dim_result}
    Suppose $\{(X_i, Y_i)\}_{i=1}^{n+1}$ are independent and identically distributed. Let $\mathcal{F} = \{\Phi(\cdot)^\top\beta : \beta \in \sR^d\}$ denote the class of linear functions over the basis $\Phi: \mathcal{X} \to \mathbb{R}^d$. If the distribution of $s \mid X$ is continuous, then for all $f \in \mathcal{F}$, we have 
    \[
    \left|\E[f(X_{n+1}) \cdot (\vone\{Y_{n+1} \in \gC(X_{n+1})\} - (1-\alpha))] \right| \leq \frac{d}{n+1}\E \left [\max_{1 \leq i \leq n+1}|f(X_{i})| \right ].
    \]   


\end{theorem}

To achieve our coverage objective (\ref{eq:V_cond_cov}), we now define a function class $\gF$ that depends on $V$. 
\begin{corollary} 
\label{corollary1}
Let $\mathcal{F} = \{\sum_{I_1 \in \mathcal{I}_1} \beta_{1_{I_1}} \vone\{\mbox{\textup{\textsf{Conf}}}(x)\in I_1\} + \sum_{I_2 \in \mathcal{I}_2} \beta_{2_{I_2}} \vone\{\mbox{\textup{\textsf{Trust}}}(x) \in I_2\}: \beta_1 \in \R^{|\mathcal{I}_1|}, \beta_2 \in \R^{|\mathcal{I}_2|}\}$ for some finite collection of sub-intervals $\mathcal{I}_1, \mathcal{I}_2$. Then,
\[
\P(Y_{n + 1} \in \gC(X_{n + 1}) \mid \mbox{\textup{\textsf{Conf}}}(X_{n+1})\in I_1, \mbox{\textup{\textsf{Trust}}}(X_{n+1}) \in I_2) \geq 1 - \alpha.
\]
If we randomize the non-conformity scores $s$, we have an upper bound on coverage,
\begin{align}
    \nonumber
    \P(Y_{n + 1} \in \gC(X_{n + 1}) \mid \mbox{\textup{\textsf{Conf}}}(X_{n+1})\in I_1,\, & \mbox{\textup{\textsf{Trust}}}(X_{n+1}) \in I_2) \leq 1 - \alpha \,+ \\
    & \frac{|\mathcal{I}_1| + |\mathcal{I}_2|}{(n + 1) \P(\mbox{\textup{\textsf{Conf}}}(X_{n+1})\in I_1,\, \mbox{\textup{\textsf{Trust}}}(X_{n+1}) \in I_2)}.
\end{align}
\end{corollary}
Note that with $\mathcal{F}$ as defined above in Corollary~\ref{corollary1}, we achieve the conditional coverage guarantee over $V$ in (\ref{eq:V_cond_cov}). However, this will be highly computationally inefficient when computing the prediction set (\ref{eq:finite_dim_pred_set}) as $\mathcal{I}_1, \mathcal{I}_2$ can be very large. In order to speed up computation we use polynomial functions of $\mbox{\textup{\textsf{Conf}}}(X)$ and $\mbox{\textup{\textsf{Trust}}}(X)$, with the intent of capturing higher-order interactions between the conformity score function and $V$.  To this end, we define $\gF$ as a function class of degree-$d$ polynomials, 
\begin{equation}
\label{eq:poly_fn_class}
    \Phi(X) = \left\{\mbox{\textup{\textsf{Conf}}}(X)^{i} \cdot \mbox{\textup{\textsf{Trust}}}(X)^{j} \mid i + j \leq d, i, j \geq 0\right\}\,;\,\gF = \left\{\Phi(\cdot)^\top\beta : \beta \in \sR^{\frac{(d+1){(d+2)}}{2}} \right\}.
\end{equation}
If we are willing to allow $\gC(X_{n + 1})$ to be randomized, we can achieve exact coverage equal to $1 - \alpha$ (stated formally in Theorem~\ref{thm:finite_dim_exact_coverage}, Appendix~\ref{sec:proofs}). However, it is not desirable to have non-deterministic prediction sets in most practical scenarios. Hence, we use a non-randomized version in our experiments, which guarantees that prediction sets have at least $1 - \alpha$ coverage over $\gF$.

\section{Experiments}

We evaluate $\conditional$ empirically on four large-scale image classification datasets and a clinical dataset in dermatology. We propose evaluation metrics that measure the gap in coverage from the desired $1 - \alpha$ level across different regions of the distribution. We also introduce a $\naive$ implementation of our proposed approach that aims to improve coverage with respect to $V$ via group-conditional coverage guarantees over non-overlapping subgroups based on $V$ (Section~\ref{sec:expt_setup}). We evaluate $\standard$, $\condnaive$, and $\conditional$ on all datasets, and show that $\conditional$ achieves the best conditional coverage performance across all settings. $\conditional$ also improves class-conditional coverage on all but one dataset.

\subsection{Evaluation metrics}
To evaluate approximate conditional coverage over test points $\{(X_{i^\prime}, Y_{i^\prime})\}_{i^\prime=1}^{N}$, we propose binning the feature space into $|B|$ bins and computing the average coverage gap (CovGap)~across~these~bins (\ref{eq:covgap}). The coverage gap measures the $l_1$ distance between the achieved coverage and the desired $1-\alpha$ level across all bins. Here, $\hat{c}_b$ denotes the mean empirical coverage in bin $b$; i.e., $\hat{c}_b = (\sum_{X_{i^\prime} \in b}  \vone\{Y_{i^\prime} \in \gC(X_{i^\prime}\}) / |b|$, where $|b|$ is the number of samples in bin $b$. This metric is inspired from the class coverage gap in~\citet{ding2023classconditional}:
\begin{equation}
\label{eq:covgap}
    \textup{CovGap} = \frac{1}{|B|} \sum_{b \in B} |\hat{c}_b - (1 - \alpha)| \times 100.
\end{equation}
For comprehensive evaluation, we measure the conditional coverage performance using CovGap over three different binning schemes:
\begin{enumerate}[1.]
    \item $\mbox{\textup{\textsf{Conf}}} \times \mbox{\textup{\textsf{Trust}}}$: We apply two-dimensional binning by splitting the samples into evenly spaced bins based on $\mbox{ \textup{\textsf{Conf}}}$ and then splitting each confidence bin into equal-size bins based on $\mbox{\textup{\textsf{Trust}}}$ score (see Figure~\ref{fig:mainfig} (bottom) for reference). 
    \item $\mbox{\textup{\textsf{Conf}}} \times \mbox{\textup{\textsf{Rank}}}$ : We apply two-dimensional binning by splitting the samples into evenly spaced bins based on $\mbox{\textup{\textsf{Conf}}}$ and then splitting each confidence bin into equal-size bins based on $\mbox{\textup{\textsf{Rank}}}$ (see Figure~\ref{fig:mainfig} (top) for reference). 
    \item Class-conditional: We split samples into bins based on their class labels, hence $|B| = |\gY|$.
\end{enumerate}

\noindent We only consider bins with a non-zero number of samples ($|b| > 0$) in our evaluation. We also report marginal coverage and average set size ($\sum_{i'=1}^{N} |\gC(X_{i^\prime})| / N$) in our results.

\subsection{Experimental setup}
\label{sec:expt_setup}

{\bf Datasets.} We perform experiments on ImageNet~\citep{russakovsky2015imagenet}, Places365~\citep{zhou2018places}, and their corresponding long-tail versions ImageNet-LT and Places365-LT~\citep{openlongtailrecognition}. ImageNet-LT and Places365-LT are constructed from the original datasets using a Pareto distribution with a power value $\alpha = 6$. Places365-LT has higher class imbalance than ImageNet-LT, defined by the number of samples in the largest class divided by the number of samples in the
smallest class. We also evaluate on the Fitzpatrick 17k dataset~\citep{groh2021evaluating} for skin disease diagnosis to study coverage of prediction sets across skin types (Section~\ref{sec:fitzpatrick}). 

\paragraph{Baselines.} Along with $\standard$ split conformal prediction, we additionally include~a~$\naive$~implementation of our approach as a baseline. $\condnaive$ uses the Mondrian conformal~prediction~procedure discussed earlier to improve coverage with respect to $V = \{ \mbox{\textup{\textsf{Conf}}}, \mbox{\textup{\textsf{Trust}}}\}$. Each individual bin $b$ in the $\mbox{\textup{\fontsize{10.5pt}{12.2pt}\textsf{Conf}}} \times \mbox{\textup{\fontsize{10.5pt}{12.2pt}\textsf{Trust}}}$ binning setting forms a group and $\condnaive$ fits a separate quantile $\hat{q}_b$ for each bin. The prediction sets for $X_{i'} \in b$ are computed as
\[
\gC(X_{i'}) = \{y : s(X_{i'}, y) \leq \hat{q}_b\}.
\]
$\condnaive$ is a competitive baseline to assess the effectiveness of our higher-dimensional~function~class~$\gF$ as it has access to the discrete binning scheme we use for evaluation.

\paragraph{Experimental details.} We use a non-randomized version of the APS score~\citep{romano2020classification} (described in Section~\ref{sec:2.1}), as our conformity score: $s(x, y) = \sum_{j=1}^{k-1} \hat{f}_{(j)}(x), \text{where} \,\, \hat{f}_{(k)}(x) = \hat{f}_y(x)$. We exclude $\hat{f}_y(x)$ to achieve smaller set sizes overall~\citep{gibbs2023conformal}. We perform an extra step of temperature scaling to rescale the probabilities following past work~\citep{angelopoulos2021uncertainty, 10.5555/3305381.3305518}. We consider $\alpha=0.1$ for all experiments to achieve a desired coverage level of 90\%. Further experimental details are provided in Appendix~\ref{sec:expt_details}.

\subsection{Results}

\begin{table}[t]
\small
\setlength{\tabcolsep}{3.8pt} 
\caption{Conditional coverage evaluation on ImageNet, ImageNet-LT, Places365, and Places365-LT. Bold indicates the best (within $\pm$0.1) coverage gap. We report standard errors in parentheses.}     
\label{table:main_results}
\centering 
\begin{tabular}{llcrcccc}
\toprule
 &  & \multicolumn{2}{c}{\textbf{Marginal}} & \multicolumn{3}{c}{\textbf{Conditional}} \\  \cmidrule(lr){3-4}\cmidrule(lr){5-7} &  & Coverage & Size &  $\mbox{\textup{\textsf{Conf}}} \times \mbox{\textup{\textsf{Trust}}}$ & $\mbox{\textup{\textsf{Conf}}} \times \mbox{\textup{\textsf{Rank}}}$   & Class-cond.  \\ &  &   &  & CovGap & CovGap &  CovGap \\
\textbf{Dataset} & \textbf{Method} &  &  &  &   &  \\
\midrule
ImageNet & \standard & 0.90 (0.00) & 4.32 (0.04) & 6.35 (0.11) & 33.12 (0.15) &  7.23 (0.04) \\
 & \textsc{conditional (naive)} & 0.93 (0.00) & 7.21 (0.05) & 8.01 (0.50) & 23.82 (0.15) & 6.68 (0.02) \\
 & \conditional & 0.90 (0.00) & 22.36 (0.83) & \textbf{4.37} (0.09) & \textbf{23.66} (0.24) &  \textbf{6.02} (0.04) \\
\midrule
ImageNet-LT & \standard & 0.89 (0.00) & 50.35 (0.02) & 4.47 (0.02) & 19.92 (0.03) &  8.32 (0.00) \\
 & \textsc{conditional (naive)} & 0.89 (0.00) & 46.63 (0.07) & 2.99 (0.01) & 18.27 (0.02) & 8.29 (0.01) \\
 & \conditional & 0.90 (0.00) & 58.64 (0.18) & \textbf{2.21} (0.01) & \textbf{17.09} (0.02) & \textbf{7.92} (0.00) \\
\midrule
Places365 & \standard & 0.90 (0.00) & 14.17 (0.07) & 5.78 (0.08) & 25.58 (0.09) &  \textbf{4.99} (0.05) \\
 & \textsc{conditional (naive)} & 0.90 (0.00) & 12.76 (0.11) & \textbf{2.40} (0.16) & 22.22 (0.13) &  5.11 (0.06) \\
 & \conditional & 0.90 (0.00) & 15.98 (0.57) & 4.38 (0.10) & \textbf{22.09} (0.12) &  \textbf{4.98} (0.05) \\
\midrule
Places365-LT & \standard & 0.90 (0.00) & 43.46 (0.10) & 5.55 (0.02) & 13.23 (0.01) & \textbf{5.34} (0.00) \\
 & \textsc{conditional (naive)} & 0.90 (0.00) & 38.54 (0.06) & 4.55 (0.02) & \textbf{11.75} (0.01) & 5.61 (0.01) \\
 & \conditional & 0.90 (0.00) & 37.07 (0.03) & \textbf{1.72} (0.00) & \textbf{11.75} (0.01) & 5.65 (0.00) \\
\bottomrule
\end{tabular}
\end{table}

Table~\ref{table:main_results} presents the CovGap of $\standard$, $\condnaive$, and $\conditional$ on all datasets. We see that while all methods achieve the desired marginal coverage level of 90\%, there is a significant difference in conditional coverage as evaluated by the coverage gap. Overall, $\conditional$ achieves the best performance (lowest CovGap) across different binning schemes in nearly all experimental settings.

Focusing our attention on the $\mbox{\textup{\textsf{Conf}}} \times \mbox{\textup{\textsf{Trust}}}$ and  $\mbox{\textup{\textsf{Conf}}} \times \mbox{\textup{\textsf{Rank}}}$ CovGap metrics, we observe that $\conditional$ significantly outperforms $\standard$ on all datasets, demonstrating improved conditional coverage. $\conditional$ also performs better or comparable to $\condnaive$ on the oracle metric $\mbox{\textup{\textsf{Conf}}} \times \mbox{\textup{\textsf{Rank}}}$  CovGap, which shows the effectiveness of our polynomial function class $\gF$ (\ref{eq:poly_fn_class}). Note that $\conditional$ has lower $\mbox{\textup{\textsf{Conf}}} \times \mbox{\textup{\textsf{Trust}}}$ coverage gap than $\condnaive$ on all but one dataset, despite the $\naive$ implementation having access to $\mbox{\textup{\textsf{Conf}}} \times \mbox{\textup{\textsf{Trust}}}$ bin information during the calibration phase. The success of both $\conditional$ and $\condnaive$ over $\standard$ empirically validates that our proposed lower-dimensional statistic $V$ is effective in improving conditional coverage. We can see this improved conditional coverage is achieved with sets that are not significantly larger than $\standard$ (and, in fact, smaller in Places365-LT). 

Further, we also evaluate class-conditional coverage gap of all methods. Despite not explicitly targeting class-conditional coverage, our method generally improves the class-conditional CovGap over $\standard$ and $\condnaive$.

\begin{figure*}[!h] 
    \centering
    \includegraphics[width=\linewidth]{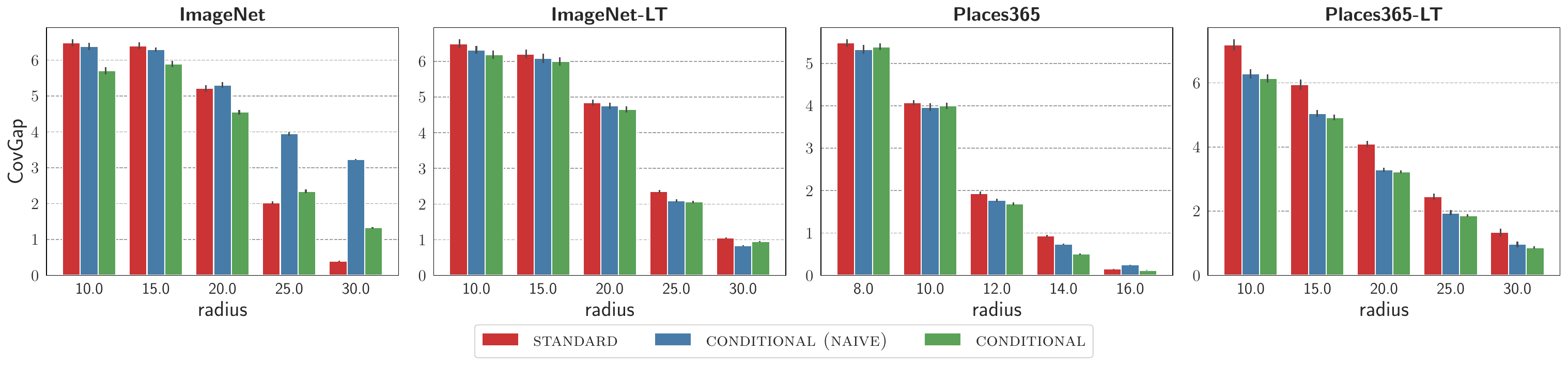} 
    \caption{\textbf{Average coverage gap between randomly sampled Euclidean balls of fixed radius.} We vary the radius on x-axis. Standard errors are reported by error bars.}
    \label{fig:euclidean_cov}
\end{figure*}
\paragraph{\bf Approximate $X$-conditional coverage.} Beyond evaluating coverage over the well-defined axes of $\mbox{\textup{\fontsize{10.5pt}{12.2pt}\textsf{Conf}}} \times \mbox{\textup{\fontsize{10.5pt}{12.2pt}\textsf{Trust}}}$ and $\mbox{\textup{\fontsize{10.5pt}{12.2pt}\textsf{Conf}}} \times \mbox{\textup{\fontsize{10.5pt}{12.2pt}\textsf{Rank}}}$, we also measure approximate conditional coverage via a notion of local coverage over $\ell_2$ balls in the feature space~\citep{barber2021limits} (Figure~\ref{fig:euclidean_cov}).  We evaluate the coverage gap between randomly sampled Euclidean balls of a fixed radius $r$, varying $r$ as shown in the figure from $[r_{min}, r_{max}]$ (see Appendix~\ref{sec:app_euclidean_cov} for further details on the choice of $r$ and simulation parameters). The CovGap in Figure~\ref{fig:euclidean_cov} shows that $\conditional$ generally performs better than $\standard$ and $\condnaive$ when $r$ is relatively small. This is an approximation of local coverage as the neighborhood is small. As $r$ approaches $r_{max}$, the coverage for individual balls approach marginal coverage, and hence the CovGap decreases and the performance of all methods is typically comparable. These results indicate that our coverage objective and method make significant progress towards improving conditional coverage in classification settings.

We compare our method with an alternative function class constructed using the top principal components of the feature layer in Appendix~\ref{sec:app_pca_expt}. Results in Table~\ref{table:app_extra_expt_pca} and Figure~\ref{fig:euclidean_cov_pca} show similar consistent improvement of $\conditional$ over this competing approach.

\begin{table}[!th]
\small
\caption{Conditional coverage evaluation on Fitzpatrick 17k. Bold indicates the best (within $\pm$0.1) coverage gap and worst-group coverage. We report standard errors in parentheses.
}     
\label{table:main_results_fitzpatrick}
\centering 
\begin{tabular}{llrccc}
\toprule
 & \multicolumn{2}{c}{\textbf{Marginal}} & \multicolumn{3}{c}{\textbf{Conditional}} \\ \cmidrule(lr){2-3}\cmidrule(lr){4-6}
 & Coverage & Size & Skin type-conditional & Worst-group & Class-conditional \\
 &  &  & CovGap & coverage & CovGap \\
\textbf{Method} &  &  &  &  &  \\
\midrule
$\standard$ & 0.90 (0.00) & 27.30 (0.12) & 1.88 (0.13) & 0.86 (0.01) & 7.69 (0.10) \\
$\groupwise$ & 0.90 (0.00) & 27.53 (0.21) & \textbf{1.76} (0.18) & 0.86 (0.01) & \textbf{7.60} (0.18) \\
$\conditional$ & 0.90 (0.00) & 30.17 (0.14) & \textbf{1.73} (0.14) & \textbf{0.87} (0.01) & \textbf{7.47} (0.13) \\
\bottomrule
\end{tabular}
\end{table}

\subsection{Fitzpatrick 17k dataset: Skin condition classification in clinical images}
\label{sec:fitzpatrick}
Fitzpatrick 17k~\citep{groh2021evaluating} is a dataset of clinical images with skin condition labels and skin type labels 1 through 6 based on the Fitzpatrick scoring system. The Fitzpatrick skin type labels are annotated by a team of humans, and a small subset of images with annotator disagreement are labeled as unknown. Higher Fitzpatrick skin type label indicates darker skin tones. The dataset has significantly fewer images of dark skin types compared to light skin, accompanied by imbalance of skin condition labels across skin types. Past work has shown there is disparity in model performance across skin tones and worse performance on uncommon diseases~\citep{doi:10.1126/sciadv.abq6147, yuksekgonul2023beyond}. Fitzpatrick 17k has 114 skin conditions (classes). Further details on the dataset and the experiment are provided in Appendix~\ref{sec:app_datasets_models}.

We evaluate how our proposed method reduces the coverage gap across skin type groups \textit{without access to type labels} (Table~\ref{table:main_results_fitzpatrick}). For this specific setup, we naturally consider skin type groups as our evaluation bins. Similar to $\condnaive$, we include an analogous $\groupwise$ baseline that has access to skin type annotations and computes an individual quantile level $\hat{q}$ for each group. We report the Skin type-conditional CovGap to measure the coverage gap between groups, along with Worst-group coverage. We see that $\conditional$ reduces the coverage gap as well as improves worst-group coverage over $\standard$. Despite having no access to group labels, $\conditional$ performs better or comparable to $\groupwise$, while having lesser variance. We also achieve lower class-conditional CovGap than $\standard$ with $\groupwise$ and $\conditional$.

\section{Discussion}
Conformal prediction guarantees exact coverage {\it marginally} across test samples in finite~samples.~However, for uncertainty sets to be truly meaningful and useful, they should provide coverage {\it conditional} on test instances where uncertainty is higher or where the model is more likely to err. Unfortunately, achieving $X$-conditional coverage in finite samples is impossible without making~distributional~assumptions. In this paper, we propose a relaxed notion of conditional coverage that improves~the~practical utility of prediction sets by ensuring coverage where it matters most---specifically,~in~cases~where~a classifier is overconfident in its incorrect predictions. To assess a classifier's overconfidence, we use its reported confidence (softmax probabilities) in combination with a nonparametric {\it trust score} that measures its alignment with the Bayes classifier. We develop a practical variant of conformal prediction that achieves approximate conditional coverage with respect to these two variables, and demonstrate that it improves conditional coverage properties in a general sense, including subgroup-level and class-conditional coverage. By reducing the coverage gap across relevant subpopulations, our resulting prediction sets can lead to fairer and improved downstream decision-making, especially in high-stakes applications where miscoverage can be consequential.

While we show trust scores significantly improve conditional coverage in conformal prediction, they also come with limitations. Particularly, in cases where the trust scores are not a good approximate of the rank of the true class, our method may not improve conditional coverage properties over standard conformal prediction. Additionally, our function class is susceptible to computational difficulties at high polynomial degrees. Future work can explore more sophisticated function classes to achieve our proposed conditional coverage objective.

\section{Acknowledgements}
We would like to thank Anastasios Angelopoulos for insightful suggestions in the early stages of this work and helpful feedback throughout the project. We also thank Tiffany Ding and Isaac Gibbs for useful feedback on the manuscript, and Mert Yuksekgonul, Miao Xiong, and Isaac Gibbs for helpful responses to questions regarding their work. This work was funded in part by the European Union (ERC-2022-SYG-OCEAN-101071601).

\bibliography{ref}
\bibliographystyle{plainnat}

\newpage 
\appendix

\section{Proofs}
\label{sec:proofs}

\textit{Proof of Theorem 1.} See Theorem 2 in~\citet{gibbs2023conformal}.

\vspace{3.2mm}
\noindent\textit{Proof of Corollary 1.} Corollary 1 follows directly from Theorem 1 in the special case where we choose $\mathcal{F} = \{\sum_{I_1 \in \mathcal{I}_1} \beta_{1_{I_1}} \vone\{\mbox{\textup{\textsf{Conf}}}(x)\in I_1\} + \sum_{I_2 \in \mathcal{I}_2} \beta_{2_{I_2}} \vone\{\mbox{\textup{\textsf{Trust}}}(x) \in I_2\}: \beta_1 \in \R^{|\mathcal{I}_1|}, \beta_2 \in \R^{|\mathcal{I}_2|}\}$.

\vspace{3.3mm}
\begin{theorem}[Proposition 4 in~\citet{gibbs2023conformal}]
\label{thm:finite_dim_exact_coverage}
    Suppose $\{(X_i, Y_i)\}_{i=1}^{n+1}$ are independent and identically distributed. Let $\mathcal{F} = \{\Phi(\cdot)^\top\beta : \beta \in \sR^d\}$ denote the class of linear functions over the basis $\Phi: \mathcal{X} \to \mathbb{R}^d$. If we optimize the dual formulation of \textup{(\ref{eq:finite_dim_reg})} and the dual solutions are computed using an algorithm that is symmetric in the input data, then the randomized prediction set $\gC_{\textup{rand}}(X_{n + 1})$ achieves exact coverage for all $f \in \mathcal{F}$:
    \[
    \E[f(X_{n + 1}) \cdot (\vone \{Y_{n + 1} \in \gC_{\textup{rand}}(X_{n + 1})\} - (1 - \alpha))] = 0.
    \]

\end{theorem}

\noindent\textit{Proof of Theorem 2.} See Proposition 4 in~\citet{gibbs2023conformal}. We formally state this result here to show that appropriate randomization of $\gC(X_{n + 1})$ can guarantee exact coverage without the continuity assumption on $s|X$ as in Theorem 1.

\section{Experimental details}
\label{sec:expt_details}

\subsection{Experimental setup}

We set $\alpha = 0.1$ for our empirical evaluation. We run all our experiments with 10 random seeds $\{1,\dots,10\}$ and report the standard errors in our results. The randomness in our experiments is over splitting the validation set into calibration and evaluation data and fitting the temperature parameter~\citep{10.5555/3305381.3305518}.

\paragraph{Evaluation metrics.} For our conditional coverage evaluation, we use the notion of CovGap (inspired by class coverage gap in~\citet{ding2023classconditional}) to measure the coverage gap across multiple bins in the feature space. We propose three binning schemes: $\mbox{\textup{\textsf{Conf}}} \times \mbox{\textup{\textsf{Trust}}}$, $\mbox{\textup{\textsf{Conf}}} \times \mbox{\textup{\textsf{Rank}}}$, and Class-conditional CovGap. To apply two-dimensional binning, we split the samples into evenly spaced bins based on $\mbox{\textup{\textsf{Conf}}}$ and then split each confidence bin into equal-size bins based on $\mbox{\textup{\textsf{Trust}}}$ score or $\mbox{\textup{\textsf{Rank}}}$ depending on the binning scheme. For both two-dimensional binning schemes, we split the samples into 10 evenly spaced confidence bins and then 4 equal-size bins based on trust score or rank. We choose this splitting to have appreciable granularity while also ensuring most bins have sufficient number of samples in all cases. As a special case for ImageNet, we manually edit the $\mbox{\textup{\textsf{Rank}}}$ bins as $\sim75\%$ test samples have accurate predictions.

\subsection{Datasets and models}
\label{sec:app_datasets_models}

We follow the data processing steps and pretrained models used by~\citet{yuksekgonul2023beyond} in our evaluation for ImageNet, ImageNet-LT, and Places365-LT.

\paragraph{ImageNet.} ImageNet~\citep{russakovsky2015imagenet} is a large-scale image classification dataset with 1000 classes. ImageNet has roughly balanced class distributions. We use the ImageNet-1k version from Torchvision~\citep{10.1145/1873951.1874254} and the pretrained ResNet50 model. We split the validation dataset evenly into calibration and evaluation splits based on the random seed.

\paragraph{ImageNet-LT.} ImageNet-LT is the long-tailed version of ImageNet with 1000 classes~\citep{openlongtailrecognition}. ImageNet-LT was constructed from the original dataset using a Pareto distribution with a power value $\alpha = 6$, and has a maximum of 1280 images per class and minimum of 5 images per class. We use the validation split as our calibration set and use the test set for evaluation. Following~\citet{yuksekgonul2023beyond}, we use the ResNeXt50 model trained on ImageNet-LT by~\citep{zhong2021mislas} in our experiments.

\paragraph{Places365.} Places365~\citep{zhou2018places} contains 10 million images from  365 classes. For Places365, we train a ResNet152 model on our own (pretrained on ImageNet). We fine-tune on the train split of the original dataset and split the validation set evenly into calibration and test splits based on the random seed.

\paragraph{Places365-LT.} Places365-LT is the long-tailed version of Places365~\citep{openlongtailrecognition}. Places365-LT was constructed from the original dataset using a Pareto distribution with a power value $\alpha = 6$, and has a maximum of 4980 images
per class and a minimum of 5 images per class. We use the validation split as our calibration set and use the test set for evaluation. Following~\citet{yuksekgonul2023beyond}, we use the ResNet152 model trained on Places365-LT by~\citep{zhong2021mislas} in our experiments.

\paragraph{Fitzpatrick 17k.} Fitzpatrick 17k~\citep{groh2021evaluating} is a dataset of clinical images with skin condition labels and skin type labels 1 through 6 based on the Fitzpatrick scoring system, with a small subset of images labeled as `Unknown' where annotators could not determine the skin type. We use the training script provided by~\citet{groh2021evaluating} to train a ResNet18 model (pretrained on ImageNet) for 50 epochs. We randomly sample 70\% of the data for fine-tuning and split the held out data evenly into calibration and evaluation splits based on the random seed. We consider the full classification task over 114 skin condition labels. For our group-conditional coverage evaluation, we consider Fitzpatrick skin types 1 through 6 as well as the Unknown type as our subgroups.

\subsection{Details on $\gF$}
\label{sec:app_function_class}

We define our function class as $\mathcal{F} = \{\sum_{I_1 \in \mathcal{I}_1} \beta_{1_{I_1}} \vone\{\mbox{\textup{\textsf{Conf}}}(x)\in I_1\} + \sum_{I_2 \in \mathcal{I}_2} \beta_{2_{I_2}} \vone\{\mbox{\textup{\textsf{Trust}}}(x) \in I_2\}: \beta_1 \in \R^{|\mathcal{I}_1|}, \beta_2 \in \R^{|\mathcal{I}_2|}\}$. To compute $\mbox{\textup{\textsf{Trust}}}(x)$, we use features from the model's penultimate layer. Following~\citet{NEURIPS2018_7180cffd}, we skip the initial filtering step of the trust score algorithm to increase computational efficiency. For calculating the nearest neighbor distance to each class for the trust scores computation, we use IndexFlatL2 from FAISS~\citep{Johnson2017BillionScaleSS}, Meta’s open-sourced GPU-accelerated library for efficient similarity search. For all experiments, we run $\conditional$ with polynomial degree $d=5$. We study the effect of $d$ in Appendix~\ref{sec:app_degree}. We note that with $d > 5$, prediction set construction considerably slows down; hence, we do not report results for polynomial degree values $d > 5$. The prediction sets in~\citet{gibbs2023conformal} are constructed by updating the quantile fit for each test point using an iterative algorithm, where the cost per iteration is $O(nd^2)$ (with a slight abuse of notation, $d$ refers to the dimension of the function class $\gF$ here). The number of iterations tends to be small in practice.

\subsection{Details on approximate $X$-conditional coverage evaluation}
\label{sec:app_euclidean_cov}

We share details regarding the experimental setup for evaluating approximate $X$-conditional coverage (Figure~\ref{fig:euclidean_cov}). In this experiment, we evaluate the coverage gap between randomly sampled $\ell_2$ balls in the feature space. We compute the Euclidean distance between features $X_i$ and $X_j$ from the model's penultimate layer for samples $i, j$. We vary the radius $r$ of the $\ell_2$ balls, where $r$ is evenly spaced 
in the interval $[r_{min}, r_{max}]$. $r_{min}$ and $r_{max}$ are estimated as the minimum and 90th percentile of the distribution of distances between randomly sampled pairs of points for each dataset, respectively. For a fixed $r$, we randomly sample 100 test points and find Euclidean balls with radius $r$ around the test point such that every ball should have at least 10 neighboring points. Then, we compute the coverage gap over these regions using Eq.~\ref{eq:covgap}. We perform 100 trials of this procedure for each $r$ for all datasets, and report the standard errors by error bars.

\section{Additional experimental results}
\label{sec:app_expts}

\subsection{Miscoverage patterns in $\standard$ and $\conditional$}

\begin{figure}[!h] 
    \centering
    \includegraphics[width=1\linewidth]{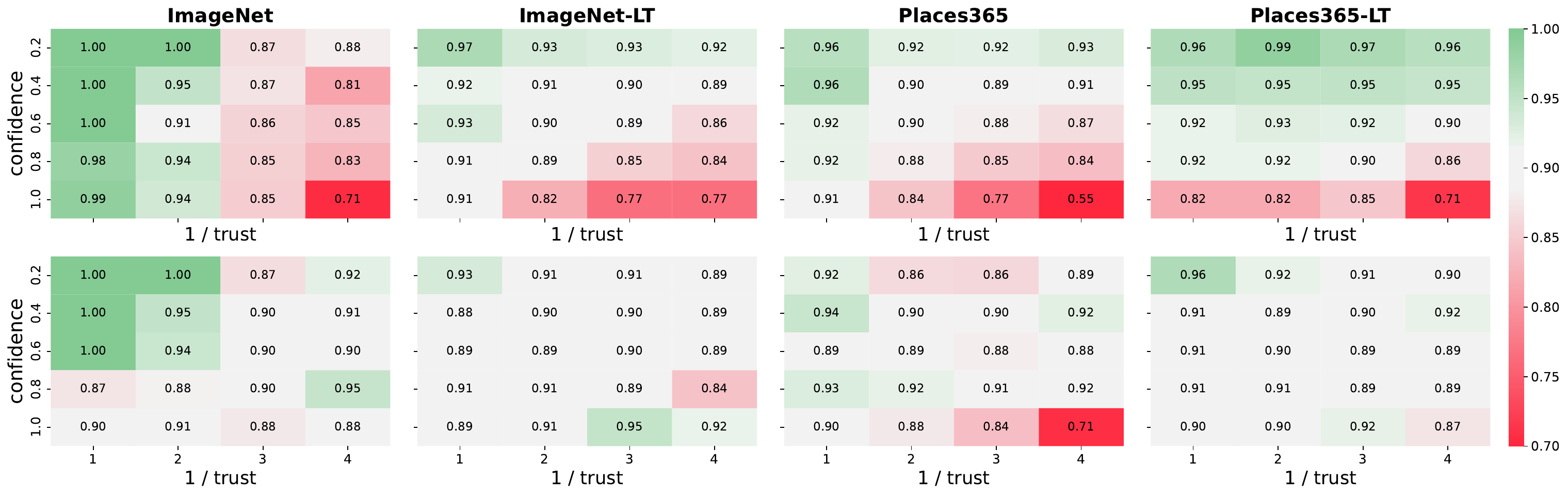} 
    \caption{Conditional coverage of $\standard$ (top) and $\conditional$ (bottom) over regions of the feature space binned by model confidence and \textit{trust score} ($\mbox{\textup{\fontsize{10.5pt}{12.2pt}\textsf{Conf}}} \times \mbox{\textup{\fontsize{10.5pt}{12.2pt}\textsf{Trust}}}$).}
    \label{fig:app_heatmap_trust}
\end{figure}

\begin{figure}[!h] 
    \centering
    \includegraphics[width=1\linewidth]{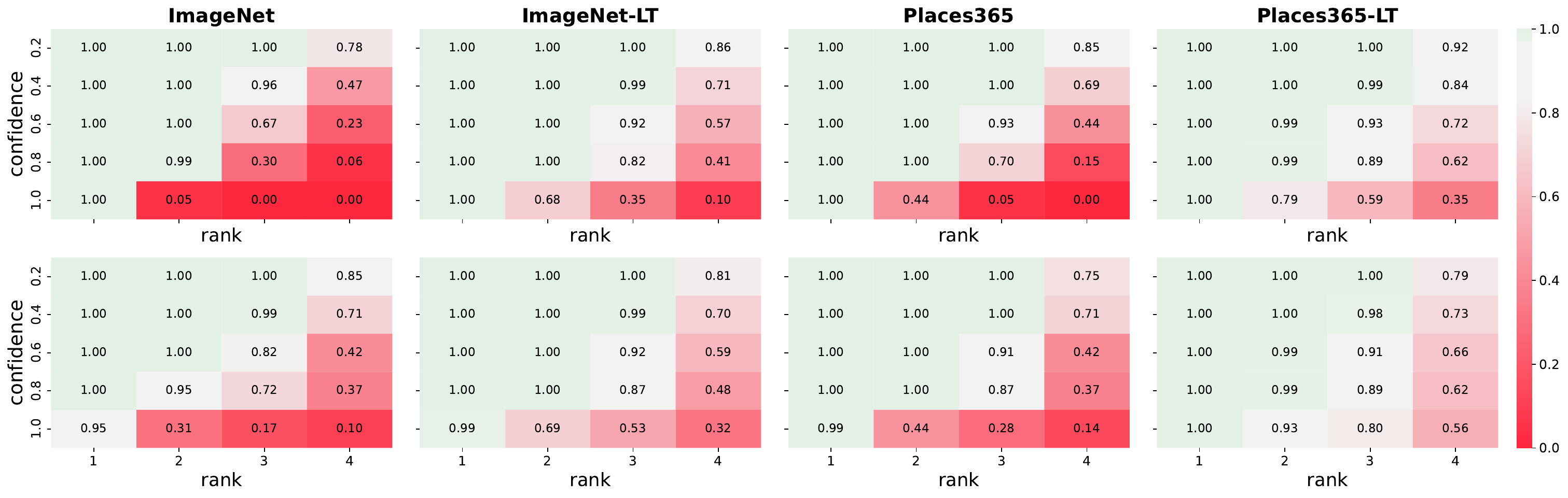} 
    \caption{Conditional coverage of $\standard$ (top) and $\conditional$ (bottom) over regions of the feature space binned by model confidence and rank of the true class ($\mbox{\textup{\fontsize{10.5pt}{12.2pt}\textsf{Conf}}} \times \mbox{\textup{\fontsize{10.5pt}{12.2pt}\textsf{Rank}}}$).}
    \label{fig:app_heatmap_rank}
\end{figure}

To demonstrate the improvement in conditional coverage over standard conformal prediction, we show the miscoverage patterns as in Figure~\ref{fig:mainfig} for $\conditional$ along with $\standard$ (Figures~\ref{fig:app_heatmap_trust},~\ref{fig:app_heatmap_rank}). Figure~\ref{fig:app_heatmap_trust} demonstrates the effectiveness of our function class $\gF$, as we see that coverage over all bins approaches the desired level of $1 - \alpha = 0.9$. From Figure~\ref{fig:app_heatmap_rank}, we can see that $\conditional$ typically improves coverage over the severly under-covered bins in $\mbox{\textup{\textsf{Conf}}} \times \mbox{\textup{\textsf{Rank}}}$ for all datasets. 
\vspace{1mm}

\subsection{Effect of $d$}
\label{sec:app_degree}

\begin{figure}[!ht]
\centering
   \includegraphics[width=0.95\linewidth]{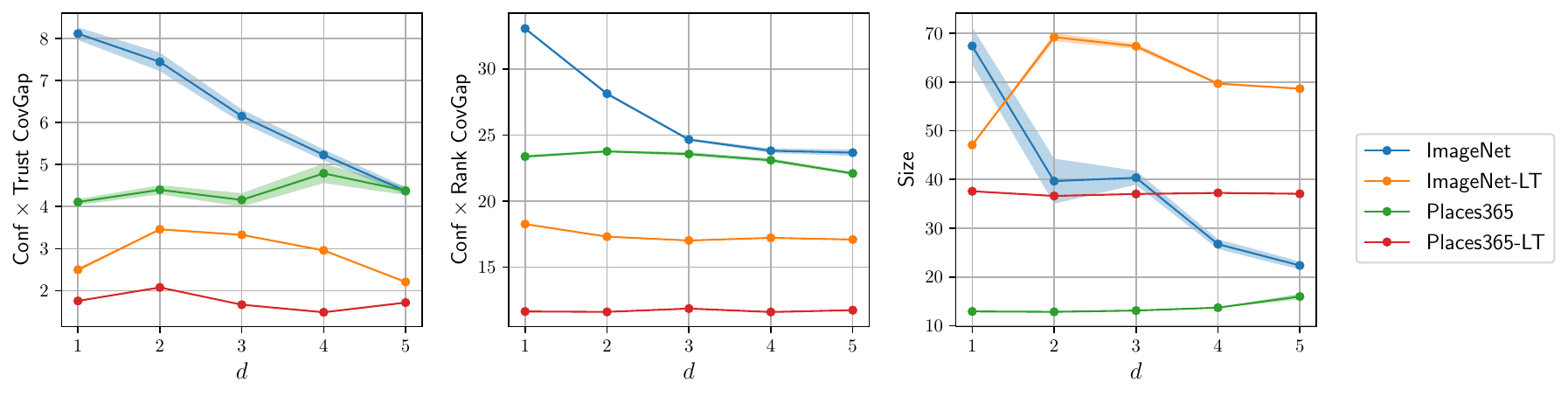}
    \caption{Effect of polynomial degree $d$ on CovGap and Size.}
    \label{fig:effect_of_degree}
\end{figure}

We study the effect of polynomial degree $d$ in function class $\gF$ (\ref{eq:poly_fn_class}) on CovGap and average set size in Figure~\ref{fig:effect_of_degree}. We can see that the choice of $d$ is not a trivial one, and different datasets may have different optimal values for $d$. The polynomial function class offers greater flexibility in capturing interactions between the conformity score function and $V$ through this choice, compared to the $\naive$ baseline. Specifically for ImageNet, it is interesting to note that the CovGap improves as we increase $d$, and the set sizes also shrink on average.

\subsection{Trust-Rank correlation}
\label{sec:app_trust_rank}

\begin{figure}[!h] 
    \centering
    \includegraphics[width=1\linewidth]{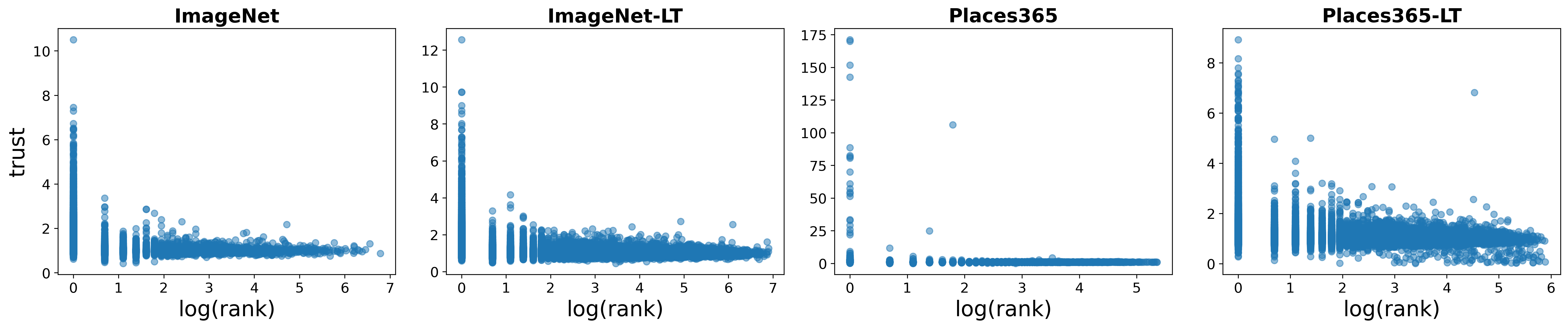} 
    \caption{Relationship between trust score ($\mbox{\textup{\small\textsf{Trust}}}$) and $\log$(rank) ($\mbox{\textup{\small\textsf{Rank}}}$).}
    \label{fig:trust_rank_corr}
\end{figure}

Here we empirically study the relationship between trust score ($\mbox{\textup{\textsf{Trust}}}$) and rank of the true class ($\mbox{\textup{\textsf{Rank}}}$). We compute the Pearson and Spearman's rank correlation coefficients with $p$-values for $\mbox{\textup{\textsf{Trust}}}$ and $\mbox{\textup{\textsf{Rank}}}$ on test samples in all datasets (Table~\ref{tab:correlation_results}). The correlation coefficients and $p$-values indicate a statistically significant negative correlation between $\mbox{\textup{\textsf{Trust}}}$ and $\mbox{\textup{\textsf{Rank}}}$. We also plot the relationship between trust score and $\log$(rank) in Figure~\ref{fig:trust_rank_corr}. This shows that lower (better) rank values generally correspond to higher trust scores on average, whereas higher (worse) rank values generally correspond to lower trust scores.

\begin{table}[!ht]
\small
\centering
\caption{Pearson and Spearman correlation coefficients with $p$-values for $\mbox{\textup{\small\textsf{Trust}}}$ and $\mbox{\textup{\small\textsf{Rank}}}$.}
\begin{tabular}{lcccccc}
\toprule
 & \multicolumn{2}{c}{\textbf{Pearson}} & \multicolumn{2}{c}{\textbf{Spearman}} \\
\cmidrule(lr){2-3} \cmidrule(lr){4-5}
& \textbf{r} & \textbf{$\bm{p}$-value} & \textbf{r} & \textbf{$\bm{p}$-value} \\
\textbf{Dataset} & & & &\\
\midrule
ImageNet & -0.09  & $< 0.001$ & -0.53  & $< 0.001$ \\
ImageNet-LT & -0.13  & $< 0.001$     & -0.47  & $< 0.001$   \\
Places365 & -0.02 & 0.004      & -0.43 & $< 0.001$      \\
Places365-LT & -0.14  & $< 0.001$       & -0.29 & $< 0.001$       \\
\bottomrule
\end{tabular}
\label{tab:correlation_results}
\end{table}

\newpage
\subsection{Principal components of feature layer as $\mathcal{F}$}
\label{sec:app_pca_expt}

\begin{figure*}[!h] 
    \centering
    \includegraphics[width=\linewidth]{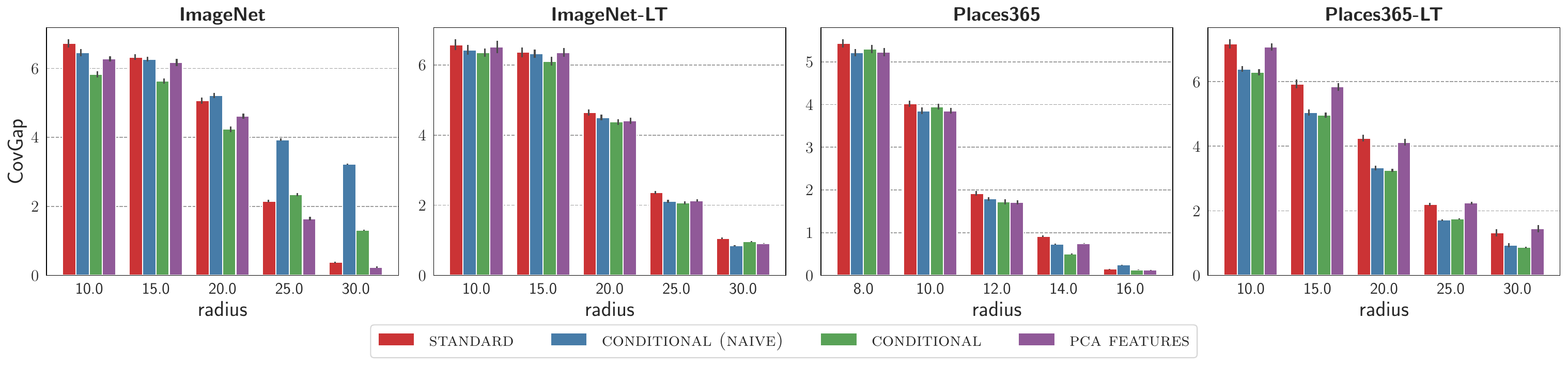} 
    \caption{\textbf{Average coverage gap between randomly sampled Euclidean balls of fixed radius.} We vary the radius on x-axis. Standard errors are reported by error bars.}
    \label{fig:euclidean_cov_pca}
\end{figure*}

We construct an alternative function $\gF$ class using the top principal components of the feature layer as input. We choose the number of principal components as 20, with an added intercept term to achieve marginal coverage. We make this choice considering the computational constraints of the conditional conformal procedure in~\citet{gibbs2023conformal} at the scale of our datasets (discussed in Appendix~\ref{sec:app_function_class}). We compare the performance of this approach with our method over all evaluation metrics. Conditional coverage evaluation in Table~\ref{table:app_extra_expt_pca} shows that $\conditional$ outperforms this function class on average over all datasets. Approximate $X$-conditional coverage evaluation (Figure~\ref{fig:euclidean_cov_pca}) also shows that our method consistently outperforms this function class on all but one dataset.

\begin{table}[!h]
\small
\setlength{\tabcolsep}{3.8pt} 
\caption{Conditional coverage evaluation on ImageNet, ImageNet-LT, Places365, and Places365-LT. Bold indicates the best (within $\pm$0.1) coverage gap. We report standard errors in parentheses.}     
\label{table:app_extra_expt_pca}
\centering 
\begin{tabular}{llcrcccc}
\toprule
 &  & \multicolumn{2}{c}{\textbf{Marginal}} & \multicolumn{3}{c}{\textbf{Conditional}} \\  \cmidrule(lr){3-4}\cmidrule(lr){5-7} &  & Coverage & Size &  $\mbox{\textup{\textsf{Conf}}} \times \mbox{\textup{\textsf{Trust}}}$ & $\mbox{\textup{\textsf{Conf}}} \times \mbox{\textup{\textsf{Rank}}}$  & Class-cond.  \\ &  &   &  & CovGap & CovGap &  CovGap \\
\textbf{Dataset} & \textbf{Method} &  &  &  &   &  \\
\midrule
ImageNet & \standard & 0.90 (0.00) & 4.32 (0.04) & 6.35 (0.11) & 33.12 (0.15) &  7.23 (0.04) \\
 & \textsc{conditional (naive)} & 0.93 (0.00) & 7.21 (0.05) & 8.01 (0.50) & 23.82 (0.15) & 6.68 (0.02) \\
 & \conditional & 0.90 (0.00) & 22.36 (0.83) & \textbf{4.37} (0.09) & \textbf{23.66} (0.24) &  \textbf{6.02} (0.04) \\
 & \textsc{pca features} & 0.90 (0.00) & 6.54 (0.07) & 6.23 (0.09) & 31.27 (0.14) & 6.92 (0.04) \\
\midrule
ImageNet-LT & \standard & 0.89 (0.00) & 50.35 (0.02) & 4.47 (0.02) & 19.92 (0.03) &  8.32 (0.00) \\
 & \textsc{conditional (naive)} & 0.89 (0.00) & 46.63 (0.07) & 2.99 (0.01) & 18.27 (0.02) & 8.29 (0.01) \\
 & \conditional & 0.90 (0.00) & 58.64 (0.18) & \textbf{2.21} (0.01) & \textbf{17.09} (0.02) & \textbf{7.92} (0.00) \\
 & \textsc{pca features} & 0.89 (0.00) & 53.59 (0.05) & 4.78 (0.01) & 20.13 (0.04) & 8.17 (0.01) \\
\midrule
Places365 & \standard & 0.90 (0.00) & 14.17 (0.07) & 5.78 (0.08) & 25.58 (0.09) &  \textbf{4.99} (0.05) \\
 & \textsc{conditional (naive)} & 0.90 (0.00) & 12.76 (0.11) & \textbf{2.40} (0.16) & 22.22 (0.13) &  5.11 (0.06) \\
 & \conditional & 0.90 (0.00) & 15.98 (0.57) & 4.38 (0.10) & \textbf{22.09} (0.12) &  \textbf{4.98} (0.05) \\
 & \textsc{pca features} & 0.90 (0.00) & 13.89 (0.07) & 5.34 (0.07) & 25.17 (0.12) & \textbf{4.96} (0.04)\\
\midrule
Places365-LT & \standard & 0.90 (0.00) & 43.46 (0.10) & 5.55 (0.02) & 13.23 (0.01) & \textbf{5.34} (0.00) \\
 & \textsc{conditional (naive)} & 0.90 (0.00) & 38.54 (0.06) & 4.55 (0.02) & \textbf{11.75} (0.01) & 5.61 (0.01) \\
 & \conditional & 0.90 (0.00) & 37.07 (0.03) & \textbf{1.72} (0.00) & \textbf{11.75} (0.01) & 5.65 (0.00) \\
 & \textsc{pca features} & 0.90 (0.00) & 45.03 (0.13) & 5.37 (0.03) & 12.89 (0.01) & \textbf{5.35 }(0.01)\\
\bottomrule
\end{tabular}
\end{table}

\newpage
\subsection{Conditional coverage evaluation using different score functions}

We motivate our method using the \textit{Adaptive Prediction Sets} (APS) algorithm~\citep{romano2020classification} designed to improve $X$-conditional coverage in classification settings. \textit{Regularized Adaptive Prediction Sets} (RAPS)~\citep{angelopoulos2021uncertainty} is a regularized version of APS that produces smaller sets on average. In Table~\ref{table:full_appendix_results}, we report the conditional coverage evaluation metrics on all datasets using APS, RAPS, and the simple softmax-based score described in Section~\ref{sec:2.1}. We note that our method can also improve conditional coverage properties using other score functions, especially in terms of class-conditional coverage gap.
\vspace{4mm}

\begin{table}[!ht]
\scriptsize
\setlength{\tabcolsep}{4.2pt} 
\caption{Conditional coverage evaluation on ImageNet, ImageNet-LT, Places365, and Places365-LT with different conformity score functions. Bold indicates the best (within $\pm$0.1) coverage gap. We report standard errors in parentheses.}     
\label{table:full_appendix_results}
\centering 
\begin{tabular}{lllcrccc}
\toprule
 &  & & \multicolumn{2}{c}{\textbf{Marginal}} & \multicolumn{3}{c}{\textbf{Conditional}} \\  \cmidrule(lr){4-5}\cmidrule(lr){6-8} 
 &  &  & Coverage & Size & $\mbox{\textup{\textsf{Conf}}} \times \mbox{\textup{\textsf{Trust}}}$ & $\mbox{\textup{\textsf{Conf}}} \times \mbox{\textup{\textsf{Rank}}}$ & Class-cond.  \\ &  &   &  &  & CovGap & CovGap &  CovGap \\
\textbf{Dataset} & \textbf{Score func.} & \textbf{Method} &  &  &  &   &  \\
\midrule
ImageNet & softmax & $\standard$ & 0.90 (0.00) & 2.12 (0.01) & 10.32 (0.15) & 32.25 (0.12) & 7.41 (0.04) \\
 &  & $\textsc{conditional (naive)}$ & 0.90 (0.00) & 4.61 (0.06) & 8.48 (0.62) & 24.18 (0.12) & 6.28 (0.04) \\
 &  & $\conditional$ & 0.90 (0.00) & 98.36 (1.84) & \textbf{5.94} (0.26) & \textbf{23.89} (0.17) & \textbf{6.09} (0.04) \\
\cline{3-8}
 & APS & $\standard$ & 0.90 (0.00) & 4.32 (0.04) & 6.35 (0.11) & 33.12 (0.15) & 7.23 (0.04) \\
 &  & $\textsc{conditional (naive)}$ & 0.93 (0.00) & 7.21 (0.05) & 8.01 (0.50) & 23.82 (0.15) & 6.68 (0.02) \\
 &  & $\conditional$ & 0.90 (0.00) & 22.36 (0.83) & \textbf{4.37} (0.09) & \textbf{23.66} (0.24) & \textbf{6.02} (0.04) \\
\cline{3-8}
 & RAPS & $\standard$ & 0.90 (0.00) & 3.03 (0.02) & 7.03 (0.08) & 33.12 (0.18) & 7.27 (0.04) \\
 &  & $\textsc{conditional (naive)}$ & 0.93 (0.00) & 5.13 (0.04) & 7.99 (0.50) & 24.04 (0.15) & 6.73 (0.01) \\
 &  & $\conditional$ & 0.90 (0.00) & 5.82 (0.06) & \textbf{4.45} (0.19) & \textbf{23.76} (0.23) & \textbf{6.07} (0.02) \\
\midrule ImageNet-LT & softmax & $\standard$ & 0.90 (0.00) & 24.54 (0.02) & 4.25 (0.02) & \textbf{17.68} (0.02) & 9.45 (0.01) \\
 &  & $\textsc{conditional (naive)}$ & 0.89 (0.00) & 33.04 (0.04) & \textbf{2.86} (0.00) & 19.86 (0.01) & 8.37 (0.00) \\
 &  & $\conditional$ & 0.90 (0.00) & 228.82 (1.44) & 6.82 (0.01) & 18.04 (0.01) & \textbf{7.68} (0.01) \\
\cline{3-8}
 & APS & $\standard$ & 0.89 (0.00) & 50.35 (0.02) & 4.47 (0.02) & 19.92 (0.03) & 8.32 (0.00) \\
 &  & $\textsc{conditional (naive)}$ & 0.89 (0.00) & 46.63 (0.07) & 2.99 (0.01) & 18.27 (0.02) & 8.29 (0.01) \\
 &  & $\conditional$ & 0.90 (0.00) & 58.64 (0.18) & \textbf{2.21} (0.01) & \textbf{17.09} (0.02) & \textbf{7.92} (0.00) \\
\cline{3-8}
 & RAPS & $\standard$ & 0.89 (0.00) & 25.79 (0.08) & 5.30 (0.02) & \textbf{16.95} (0.03) & 9.55 (0.00) \\
 &  & $\textsc{conditional (naive)}$ & 0.89 (0.00) & 33.82 (0.00) & 2.98 (0.01) & 19.95 (0.02) & 8.50 (0.00) \\
 &  & $\conditional$ & 0.90 (0.00) & 35.04 (0.02) & \textbf{1.72} (0.01) & 18.73 (0.02) & \textbf{8.15} (0.01) \\
\midrule Places365 & softmax & $\standard$ & 0.90 (0.00) & 9.38 (0.03) & 3.18 (0.06) & \textbf{20.21} (0.13) & 5.30 (0.05) \\
 &  & $\textsc{conditional (naive)}$ & 0.90 (0.00) & 10.37 (0.06) & \textbf{2.39} (0.14) & 22.36 (0.10) & 5.04 (0.05) \\
 &  & $\conditional$ & 0.90 (0.00) & 43.96 (1.68) & 6.64 (0.21) & 23.15 (0.12) & \textbf{4.86} (0.04) \\
\cline{3-8}
 & APS & $\standard$ & 0.90 (0.00) & 14.17 (0.07) & 5.78 (0.08) & 25.58 (0.09) & \textbf{4.99} (0.05) \\
 &  & $\textsc{conditional (naive)}$ & 0.90 (0.00) & 12.76 (0.11) & \textbf{2.40} (0.16) & 22.22 (0.13) & 5.11 (0.06) \\
 &  & $\conditional$ & 0.90 (0.00) & 15.98 (0.57) & 4.38 (0.10) & \textbf{22.09} (0.12) & \textbf{4.98} (0.05) \\
\cline{3-8}
 & RAPS & $\standard$ & 0.90 (0.00) & 10.24 (0.05) & 3.60 (0.08) & \textbf{21.43} (0.13) & 5.21 (0.04) \\
 &  & $\textsc{conditional (naive)}$ & 0.90 (0.00) & 10.93 (0.09) & \textbf{2.44} (0.14) & 22.38 (0.09) & 5.05 (0.05) \\
 &  & $\conditional$ & 0.90 (0.00) & 11.48 (0.05) & 3.99 (0.09) & 22.19 (0.10) & \textbf{4.99} (0.04) \\
\midrule Places365-LT & softmax & $\standard$ & 0.90 (0.00) & 24.98 (0.05) & 2.59 (0.00) & 14.88 (0.01) & \textbf{6.24} (0.00) \\
 &  & $\textsc{conditional (naive)}$ & 0.90 (0.00) & 26.45 (0.03) & 4.62 (0.01) & 14.98 (0.02) & 6.36 (0.01) \\
 &  & $\conditional$ & 0.90 (0.00) & 28.35 (0.25) & \textbf{2.02} (0.01) & \textbf{14.44} (0.01) & \textbf{6.23} (0.01) \\
\cline{3-8}
 & APS & $\standard$ & 0.90 (0.00) & 43.46 (0.10) & 5.55 (0.02) & 13.23 (0.01) & \textbf{5.34} (0.00) \\
 &  & $\textsc{conditional (naive)}$ & 0.90 (0.00) & 38.54 (0.06) & 4.55 (0.02) & \textbf{11.75} (0.01) & 5.61 (0.01) \\
 &  & $\conditional$ & 0.90 (0.00) & 37.07 (0.03) & \textbf{1.72} (0.00) & \textbf{11.75} (0.01) & 5.65 (0.00) \\
\cline{3-8}
 & RAPS & $\standard$ & 0.90 (0.00) & 23.46 (0.10) & 3.00 (0.01) & 15.84 (0.11) & 6.47 (0.00) \\
 &  & $\textsc{conditional (naive)}$ & 0.90 (0.00) & 26.23 (0.04) & 4.31 (0.01) & 14.91 (0.05) & 6.44 (0.01) \\
 &  & $\conditional$ & 0.90 (0.00) & 25.92 (0.05) & \textbf{1.51} (0.01) & \textbf{14.81} (0.05) & \textbf{6.37} (0.01) \\
\bottomrule
\end{tabular}
\end{table}

\end{document}